%% file: main.tex
\documentclass[sigconf]{acmart}

\usepackage{subfig}
\input{dlab_macros}

\newcommand{\WP}{Wikipedia\xspace}
\newcommand{\cl}{crosslingual\xspace}
\newcommand{\githubURL}{\url{https://github.com/epfl-dlab/WikiPDA}}
\newcommand{\BibTeX}{B\textsc{ib}\TeX}
\AtBeginDocument{%
  \providecommand\BibTeX{{%
    \normalfont B\kern-0.5em{\scshape i\kern-0.25em b}\kern-0.8em\TeX}}}

\copyrightyear{2021}
\acmYear{2021}
\setcopyright{iw3c2w3}
\acmConference[WWW '21]{Proceedings of The Web Conference 2021}{April 19--23, 2021}{Ljubljana, Slovenia}
\acmBooktitle{Proceedings of The Web Conference 2021 (WWW '21), April 19--23, 2021, Ljubljana, Slovenia}
\acmPrice{}
\acmDOI{10.1145/3442381.3449805}
\acmISBN{978-1-4503-8312-7/21/04}

\begin{document}

\title{Crosslingual Topic Modeling with WikiPDA}


\author{Tiziano Piccardi}
\affiliation{%
  \institution{EPFL}
}
\email{tiziano.piccardi@epfl.ch}

\author{Robert West}
\authornote{Robert West is a Wikimedia Foundation Research Fellow.}
\affiliation{%
  \institution{EPFL}
}
\email{robert.west@epfl.ch}

\begin{abstract}
We present \textit{Wikipedia-based Polyglot Dirichlet Allocation (WikiPDA),} a crosslingual topic model that learns to represent Wikipedia articles written in any language as distributions over a common set of language-independent topics.
It leverages the fact that Wikipedia articles link to each other and are mapped to concepts in the Wikidata knowledge base, such that, when represented as bags of links, articles are inherently language-independent.
WikiPDA works in two steps, by first densifying bags of links using matrix completion and then training a standard monolingual topic model.
A human evaluation shows that WikiPDA produces more coherent topics than monolingual text-based latent Dirichlet allocation (LDA), thus offering crosslinguality at no cost.
We demonstrate WikiPDA's utility in two applications: a study of topical biases in 28 Wikipedia language editions, and crosslingual supervised document classification.
Finally, we highlight WikiPDA's capacity for zero-shot language transfer, where a model is reused for new languages without any fine-tuning.
Researchers can benefit from WikiPDA as a practical tool for studying Wikipedia's content across its 299 language editions in interpretable ways, via an easy-to-use library publicly available at \githubURL.
\end{abstract}

\maketitle

\section{Introduction}
\label{sec:Introduction}



With 53 million articles written in 299 languages, \WP is the largest encyclopedia in history.
To leverage and analyze individual language editions, researchers have successfully used topic models~\cite{Singer2017-dt}.
The goal of this paper is to move beyond individual language editions and develop a topic model that works for all language editions jointly.
Our method, \textit{Wikipedia\hyp based Polyglot Dirichlet Allocation (WikiPDA),} learns to represent articles written in any language in terms of language\hyp independent, interpretable semantic topics.
This way, articles that cannot be directly compared in terms of the words they contain (as the words are from different vocabularies) can nevertheless be compared in terms of their topics.


Such a model is tremendously useful in practice.
With close to a billion daily page views, \WP plays an important role in everyday life,
and it is equally important as a dataset and object of study for researchers across domains:
Google Scholar returns about 2 million publications for the query ``Wikipedia''.
Although English is but one of 299 language editions, it is currently by far the most studied by researchers, to an extent that goes well beyond what can be justified by size alone.%
\footnote{For instance, with 6 million articles, the English edition is 5 times as large as the Vietnamese one, whereas Google Scholar returns over 300 times as many results for ``en.wikipedia.org'' (387K) as for ``vi.wikipedia.org'' (1,250).}
The scarcity of easy-to-use \cl topic models contributes to this skew, affecting even those studies that go beyond English; \eg, it kept \citet{lemmerich2019world}, who compared the usage of 14 language editions via survey data and browsing logs, from quantifying differences in users' topical interests across languages.

\begin{figure}
\includegraphics[width=\columnwidth]{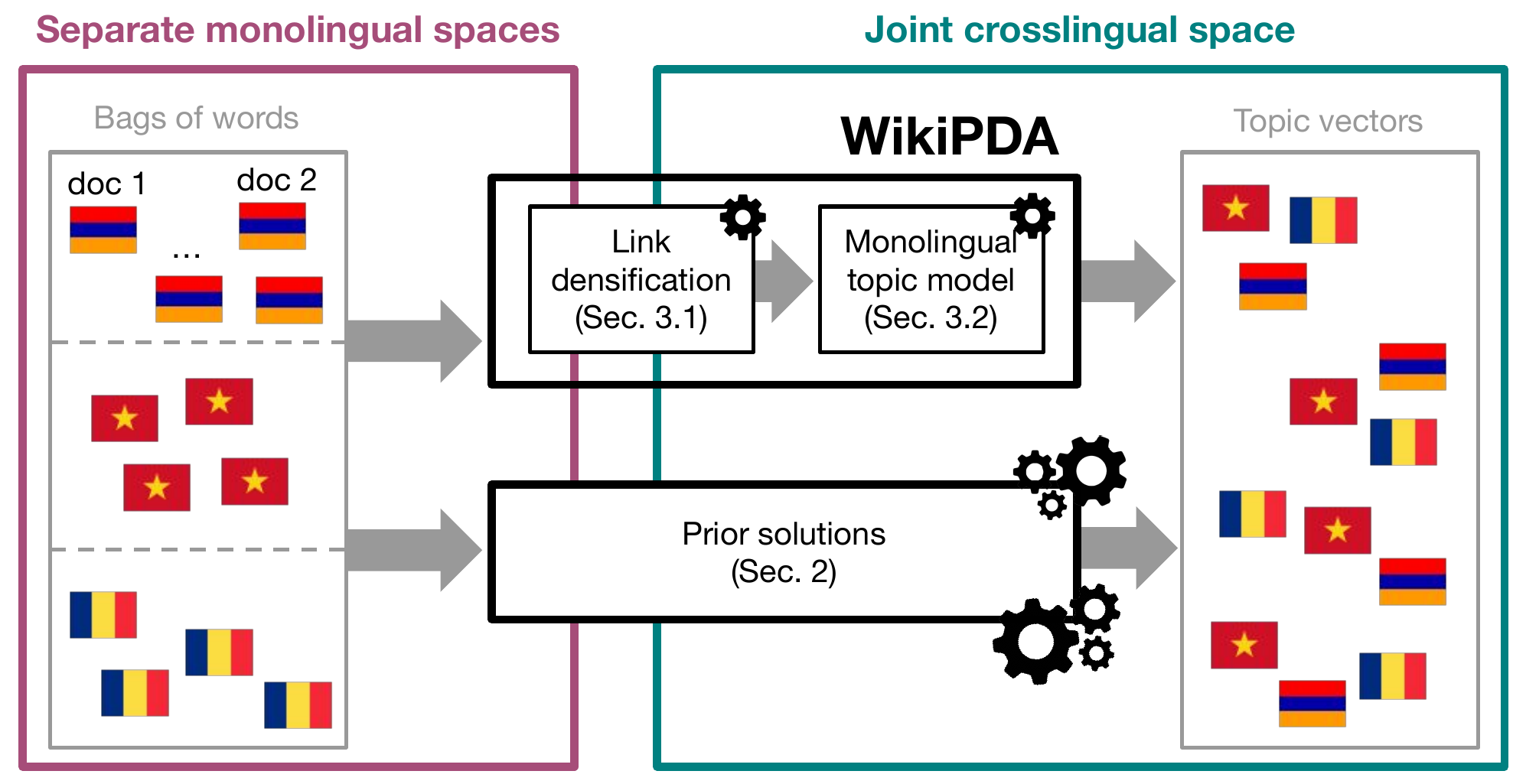}
\caption{
Overview of Wikipedia-based Polyglot Dirichlet Allocation (WikiPDA).
Whereas prior \cl topic models typically learned to directly map from separate monolingual bag-of-word spaces to a joint \cl topic\hyp vector space,
WikiPDA proceeds in two steps:
In the first step, language\hyp specific bags of words are mapped to language\hyp independent bags of out-links, using the fact that each language\hyp specific \WP article (and thus each out-link) corresponds to one language\hyp independent concept in the Wikidata knowledge base.
In the second step, the language\hyp independent bags of out-links are fed to a vetted, powerful monolingual topic model such as LDA.
}
\label{fig:overview}
\end{figure}


Although each language on its own can be readily handled via standard topic models, which are based on bags of words and thus straightforward to apply to any language with minimal preprocessing,
such models are insufficient for comparing content across languages because in general the topics learned for one language do not have clearly corresponding topics in the other languages.


\xhdr{Prior solutions}
To address this problem, researchers have extended monolingual topic models by mapping documents from separate monolingual spaces into a joint \cl topic space.
This paradigm has been proposed under various names (\eg,
\cl,
multilingual,
polylingual,
bilingual,
or correspondence
topic models; \cf\ \Secref{sec:Related work}),
but the basic idea is identical, namely to enhance the model by allowing for multiple languages while enforcing \cl alignment at the level of words or documents.
For instance, in document\hyp alignment models, a topic is modeled not as a single word distribution, but as a set of word distributions---one per language---, and different language versions of the same document are constrained to the same mix of topics during training.
As \WP articles are aligned across languages
via the Wikidata knowledge base,%
\footnote{For instance, the Wikidata concept Q44 corresponds to the English article \cpt{Beer}, French \cpt{Bi\`ere}, Finnish \cpt{Olu}, \etc}
\WP has served as a prominent training dataset for models based on document alignment.


\xhdrNoPeriod{Proposed solution: WikiPDA.%
\footnote{
\label{fn:github}
Code, library, models, data: \githubURL
}}
We leverage \WP's \cl article alignment from a different angle, by recognizing that \WP articles are not just plain text, but laced with links to other articles.
An article's set of outgoing links (\textit{``bag of links''}) is a concise summary of the article's key content.
Crucially, since each linked article is itself associated with a language\hyp independent Wikidata concept, bags of links immediately give rise to a \cl input representation ``for free''.
Starting from this representation, WikiPDA works in two steps, by first densifying bags of links using matrix completion and then training a standard monolingual topic model.
Whereas in previous methods, translating from mono- to crosslingual space constitutes the core computation, in WikiPDA it constitutes a mere preprocessing step
(\Figref{fig:overview}).
Put differently, whereas prior work has harnessed \WP's crosslinguality to \textit{increase model complexity,} we leverage it to \textit{decrease data complexity.}
This way, WikiPDA can leverage, as its core computation, standard monolingual topic models such as latent Dirichlet allocation (LDA)~\cite{Blei:2003:LDA:944919.944937}, which have been vetted in practice, come with implementations on all platforms, and scale to massive datasets.

A human evaluation shows that WikiPDA topics extracted jointly from 28 language editions of \WP are more coherent than those from monolingual text-based LDA, thus offering crosslinguality at no cost (\Secref{sec:evaluation}).
We demonstrate WikiPDA's practical utility in two applications (\Secref{sec:Applications}):
a topical comparison of \WP across 28 languages,
and \cl supervised document classification.
Finally, we show WikiPDA's ability to operate in the challenging zero-shot setting (\Secref{sec:Zero-shot language transfer}), where a model is applied to new languages without any fine\hyp tuning,
a powerful capacity not shared by prior \cl topic models, to the best of our knowledge.

With WikiPDA, researchers possess a new tool for studying the content of all of Wikipedia's 299 language editions in a unified framework, thus better reflecting Wikipedia's linguistic diversity. 
 

\section{Related work}
\label{sec:Related work}

Topic models \cite{blei2012probabilistic} are unsupervised machine learning techniques that represent documents as low\hyp dimensional vectors whose dimensions are interpretable as topics.
Crosslingual topic models \cite{hao2020empirical,vulic2015probabilistic} allow for documents to be written in different languages and represent them in terms of topics that are language\hyp independent.
Most \cl topic models are based on latent Dirichlet allocation (LDA) \cite{Blei:2003:LDA:944919.944937}, although some use different techniques, such as clustering based on Gaussian mixture models \cite{chan2020reproducible}.

One set of methods uses document\hyp aligned corpora, where documents that are loosely equivalent but written in different languages (\eg, \WP articles about the same concept in different languages) are grouped and constrained to the same topic distribution during training
\cite{de2009cross, Mimno:2009:PTM:1699571.1699627, Ni:2009:MMT:1526709.1526904, fukumasu2012symmetric, zhang2013cross}.

Another set of methods does not use document\hyp aligned corpora, but word alignments from bilingual dictionaries, modeling topics as distributions over \cl equivalence classes of words
\cite{jagarlamudi2010extracting, zhang2010cross,hao2018learning}.
\citet{boyd2009multilingual} require neither an aligned corpus nor a dictionary.

Document\hyp alignment\hyp based methods make effective use of large aligned corpora, but are hampered by requiring that aligned documents be about the same topics, which is frequently not the case in practice and eliminates an important use case of \cl topic models \textit{ab ovo,} namely quantifying how an identical concept is described in different languages (\cf\ \Secref{sec:Comparing Wikipedia across languages}).
Word\hyp alignment\hyp based methods, on the contrary, do not suffer from this shortcoming, but are hampered by the scarcity of multilingual dictionaries beyond two languages.

WikiPDA marries the best of both worlds by leveraging the \textit{document alignment} provided by Wikidata in the spirit of \textit{word alignment} methods:
representing articles as bags of links, rather than bags of words, may be seen as inducing a common vocabulary spanning 299 languages, without unnaturally forcing corresponding articles in different languages to have identical topic distributions.

\begin{table*}[t]
  \centering
  \caption{
  Statistics of the 28 \WP language editions used in this paper.
  }
  \begin{tabular}{ll|rrr|rrr|rrrrr}
  \multicolumn{2}{c|}{} &
  \multicolumn{3}{c|}{\textbf{Articles (in thousands)}} &
  \multicolumn{3}{c|}{\textbf{Links (in millions)\textsuperscript{\textdagger}}} & \multicolumn{5}{c}{\textbf{Disambiguation evaluation\textsuperscript{\textdagger} (\Secref{sec:evaluation: link densification})}}\\
\hline
 & & & Num.\ w/ & \% of & & & Dens. & Ambig. & \multicolumn{4}{c}{Accuracy for num.\ candidates in interval}\\
 & Language & Num. & $\geq$10 links* & total\textsuperscript{\textdagger} & Sparse & Densified & ratio\textsuperscript{\ddag} & anchors & $[1,\infty]$ & $[2,\infty]$** & $[1,10]$ & $[2,10]$**\\
\hline
ar & Arabic & 987 & 507 & 2\% & 14.6 & 49.1 & 3.4 & 48\% & 0.86 & 0.70 (0.24) & 0.88 & 0.75 (0.34)\\
ca & Catalan & 611 & 468 & 2\% & 16.2 & 71.4 & 4.4 & 46\% & 0.88 & 0.74 (0.23) & 0.90 & 0.78 (0.33)\\
cs & Czech & 410 & 357 & 1\% & 14.3 & 52.0 & 3.6 & 48\% & 0.86 & 0.71 (0.26) & 0.88 & 0.74 (0.34)\\
de & German & 2043 & 1851 & 7\% & 70.1 & 304.1 & 4.3 & 44\% & 0.86 & 0.68 (0.22) & 0.88 & 0.73 (0.33)\\
el & Greek & 164 & 117 & $<$1\% & 4.1 & 17.1 & 4.1 & 46\% & 0.87 & 0.71 (0.26) & 0.89 & 0.75 (0.33)\\
en & English & 5571 & 4511 & 18\% & 206.9 & 594.3 & 2.9 & 39\% & 0.88 & 0.68 (0.18) & 0.90 & 0.74 (0.33)\\
es & Spanish & 1461 & 1332 & 5\% & 56.8 & 179.9 & 3.2 & 37\% & 0.86 & 0.63 (0.19) & 0.89 & 0.70 (0.33)\\
fa & Persian & 674 & 341 & 1\% & 8.7 & 34.8 & 4.0 & 49\% & 0.86 & 0.71 (0.22) & 0.89 & 0.78 (0.33)\\
fi & Finnish & 451 & 348 & 1\% & 10.4 & 31.8 & 3.1 & 54\% & 0.86 & 0.75 (0.25) & 0.89 & 0.80 (0.35)\\
fr & French & 2013 & 1684 & 7\% & 81.5 & 247.3 & 3.0 & 42\% & 0.85 & 0.64 (0.19) & 0.89 & 0.73 (0.32)\\
he & Hebrew & 239 & 229 & 1\% & 12.3 & 52.7 & 4.3 & 47\% & 0.87 & 0.72 (0.25) & 0.89 & 0.76 (0.33)\\
id & Indonesian & 495 & 345 & 1\% & 8.9 & 33.8 & 3.8 & 52\% & 0.85 & 0.71 (0.23) & 0.87 & 0.76 (0.33)\\
it & Italian & 1458 & 1093 & 4\% & 54.0 & 193.6 & 3.6 & 45\% & 0.84 & 0.64 (0.20) & 0.88 & 0.73 (0.33)\\
ja & Japanese & 1097 & 1030 & 4\% & 60.4 & 80.8 & 1.3 & 44\% & 0.84 & 0.64 (0.22) & 0.87 & 0.71 (0.33)\\
ko & Korean & 418 & 307 & 1\% & 12.3 & 28.8 & 2.3 & 42\% & 0.84 & 0.63 (0.25) & 0.89 & 0.74 (0.35)\\
nl & Dutch & 1889 & 958 & 4\% & 33.2 & 116.2 & 3.5 & 55\% & 0.84 & 0.71 (0.22) & 0.87 & 0.76 (0.33)\\
pl & Polish & 1289 & 986 & 4\% & 35.3 & 105.4 & 3.0 & 43\% & 0.88 & 0.72 (0.21) & 0.90 & 0.76 (0.31)\\
pt & Portuguese & 964 & 742 & 3\% & 28.0 & 102.2 & 3.6 & 40\% & 0.86 & 0.64 (0.22) & 0.88 & 0.70 (0.33)\\
ro & Romanian & 378 & 240 & 1\% & 7.4 & 29.2 & 3.9 & 46\% & 0.90 & 0.78 (0.24) & 0.90 & 0.79 (0.32)\\
ru & Russian & 1406 & 1143 & 4\% & 47.7 & 172.9 & 3.6 & 37\% & 0.87 & 0.66 (0.21) & 0.90 & 0.72 (0.31)\\
sq & Albanian & 71 & 19 & $<$1\% & 0.7 & 2.6 & 3.8 & 53\% & 0.89 & 0.79 (0.33) & 0.89 & 0.79 (0.36)\\
sr & Serbian & 579 & 424 & 2\% & 9.7 & 46.0 & 4.7 & 50\% & 0.87 & 0.75 (0.27) & 0.89 & 0.79 (0.33)\\
sv & Swedish & 3453 & 3178 & 12\% & 59.3 & 118.8 & 2.0 & 60\% & 0.91 & 0.85 (0.26) & 0.92 & 0.87 (0.36)\\
tr & Turkish & 319 & 227 & 1\% & 7.0 & 20.8 & 3.0 & 48\% & 0.86 & 0.71 (0.24) & 0.89 & 0.78 (0.34)\\
uk & Ukrainian & 905 & 742 & 3\% & 23.0 & 80.6 & 3.5 & 45\% & 0.88 & 0.73 (0.25) & 0.90 & 0.78 (0.33)\\
vi & Vietnamese & 1218 & 543 & 2\% & 15.0 & 71.8 & 4.8 & 59\% & 0.83 & 0.72 (0.30) & 0.86 & 0.75 (0.39)\\
war & Waray & 1251 & 1142 & 4\% & 15.6 & 29.8 & 1.9 & 99\% & 0.46 & 0.46 (0.37) & 0.46 & 0.46 (0.37)\\
zh & Chinese & 1028 & 576 & 2\% & 23.4 & 31.8 & 1.4 & 54\% & 0.85 & 0.72 (0.25) & 0.88 & 0.77 (0.34)\\
\hline
& Average &1173 & 908& & 33.4& 103.5& 3.3&48\%&0.85&0.69 (0.24)&0.87&0.74 (0.33)\\
& Total & 32844& 25437& 100\%&936.8&2900.0&&&&&&\\
\hline
  \end{tabular}
  \flushleft
  *Links counted after link densification.
  \textsuperscript{\textdagger}Considering only articles with $\geq$10 links after densification.
  **Random baseline in parentheses.\\
  \textsuperscript{\ddag}Densification ratio $=\text{Densified}/\text{Sparse}$.
  \label{tab:langs}
\end{table*}


\section{WikiPDA: Wikipedia-based Polyglot Dirichlet Allocation}
\label{sec:WikiPDA}

Existing \cl topic models take a monolithic approach, mapping directly from monolingual bags of words to \cl topic distributions. WikiPDA, on the contrary, procedes sequentially (\Figref{fig:overview}), first mapping monolingual bags of words to \cl bags of links (\Secref{sec:Link densification}) and then mapping \cl bags of links to \cl topic distributions (\Secref{sec:Topic modeling}).
In what follows, we describe these two stages in turn.


\subsection{Link densification}
\label{sec:Link densification}

\WP's \textit{Manual of Style}%
\footnote{
\url{https://en.wikipedia.org/wiki/Wikipedia:Manual_of_Style/Linking}
}
asks authors to add links that aid navigation and understanding.
Key concepts are thus linked to their articles, allowing us to use bags of links, in lieu of bags of words, as concise article summaries.
Crucially, bag-of-links elements---articles---are associated with language\hyp independent Wikidata concepts, so in principle, the \cl article representations
to be fed to the downstream topic model
may be obtained simply by extracting links from articles.

In practice, however, human editors frequently fail to add all relevant links \cite{West:2009:CWH:1645953.1646093}, and they are explicitly instructed to add links parsimoniously 
(\eg, by linking only the first mention of every concept).
For topic modeling, such human\hyp centric factors are of no concern; rather, we prefer semantically complete article summaries with information about the frequency of constituent concepts.
Hence, the first phase of WikiPDA is link densification, where we link as many plain-text phrases as possible to the corresponding Wikidata concepts
(\eg, all occurrences of ``beer'', ``Beer'', ``beers'', \etc, in the article about \cpt{India pale ale} should be linked to Wikidata concept~Q44).

The difficulty arises from ambiguous phrases
(\eg, in some contexts, ``Beer'' should be linked to \cpt{Beer, Devon} [Q682112], an English village).
Disambiguating phrases to the correct Wikidata concept is the so-called ``wikification'' task, with several existing solutions \cite{mihalcea+csomai2007, Milne:2008:LLW:1458082.1458150, West:2009:CWH:1645953.1646093, noraset2014adding}, any of which could be plugged in.
Given the scale of our setting, we opted for a lightweight approach based on matrix completion:
First, given a \WP language edition, build the adjacency matrix $A$ of the hyperlink graph, where both rows and columns represent Wikidata concepts, and entry $a_{ij}$ is non-zero (details in \Secref{sec:Implementation details}) iff the article about concept $i$ contains a link to that about concept $j$.
Then, decompose
$A \approx UV^\top$
using alternating least squares \cite{Koren09matrixfactorization}, such that both $U$ and $V$ are of low rank $r$.
The rows of $U$ are latent representations of articles when serving as link sources,
and the rows of $V$, when serving as link targets,
optimized such that, for existing links $(i,j)$, we have $a_{ij} \approx u_i v_j^\top$
(where single subscripts are row indices).
For non-existing links $(i,j)$, the dot product $u_i v_j^\top$ provides a score that captures how well the new link $(i,j)$ would be in line with the existing link structure.

Thus, the scores $u_i v_j^\top$ can be used to disambiguate the plain-text phrases $p$ in article $i$: consider as the set $C_p$ of candidate targets for $p$ all articles $j$ for which $p$ occurs as an anchor at least once in the respective language edition of \WP, and select the candidate with the largest score, \ie, link the phrase $p$ in article $i$ to
$\argmax_{j \in C_p} u_i v_j^\top$.

In principle, a decomposition computed for one language can be used to disambiguate links in any other language.
For this paper, however, we computed a separate decomposition for each language,
in order to be able to model language\hyp specific patterns.


\subsection{Topic modeling}
\label{sec:Topic modeling}

The bags of links resulting from link densification can be fed to any monolingual topic model based on bags of words,
by using a vocabulary consisting of Wikidata concepts rather words,
and by using as the document corpus the union of all \WP articles pooled across all languages considered.
Concretely, we use LDA as the topic model,
but any other model based on bags of words, such as PLSA \cite{hofmann1999probabilistic}, would be compatible with our method.
As usual, the number $K$ of topics is set manually by the user.


\subsection{Implementation and corpus details}
\label{sec:Implementation details}

\xhdr{Link densification}
We considered as potential anchors for new links all 1- to 4-grams, with preference given to longer $n$-grams
(\eg, ``India pale ale'' as a whole is linked to \cpt{India pale ale}, rather than ``India'' to \cpt{India}, and ``pale ale'' to \cpt{Pale ale}).
We did not consider $n$-grams whose occurrences are linked with a probability below the threshold of 6.5\% \cite{Milne:2008:LLW:1458082.1458150}
(\eg, ``a'', ``the'', \etc),
since, like stop words, they usually do not represent semantically relevant content.

Decompositions of the adjacency matrix $A$ used rank $r=150$.
Before the decomposition, $A$'s entries were weighted in the spirit of inverse document frequency, giving more weight to links occurring in few articles:
if $i$ links to $j$, we set $a_{ij} = -\log(d_j/N)$, and $a_{ij}=0$ otherwise, where $d_j$ is
the number of articles that link to $j$,
and $N$ is the number of articles in the respective \WP \cite{Milne08aneffective}.

\xhdr{Topic modeling}
Since LDA may perform poorly with short documents \cite{Tang2014}, we removed articles with fewer than 10 links after densification.
Further, we ignored concepts appearing as links in fewer than 500 articles across all languages.

\xhdr{Corpus}
We worked with 28 language editions of \WP (details in \Tabref{tab:langs}), in their snapshots of 20~February 2020.
We work only with articles from namespace 0 (the main namespace).
Links and anchor texts were extracted from wiki markup.
Redirects were resolved.
After all preprocessing, the corpus encompassed 25M documents across all 28 languages,
with a vocabulary of 437K unique Wikidata concepts.

\xhdr{Code and model availability}
We release code and pre-trained models
(\cf\ footnote~\ref{fn:github})
for a wide range of $K$.
For $K=40$ and 100, topics were manually labeled with names.
On a single machine (48 cores, 250~GB RAM), the full pipeline for all 28 languages with fixed hyperparameters ran in under 24 hours.
As the code uses Apache Spark, parallelizing over many machines is straightforward
and would further reduce the runtime.


\section{Evaluation}
\label{sec:evaluation}

Next, we evaluate the two stages of our pipeline, link densification (\Secref{sec:evaluation: link densification}) and topic modeling (\Secref{sec:evaluation: topic modeling}).

\subsection{Link densification}
\label{sec:evaluation: link densification}

Densification increased the number of links substantially, by a factor of 3.3, to an effective
114 links per article, on average over all 28 languages
(details in \Tabref{tab:langs}).


The large fraction of ambiguous anchors (48\%) underlines the importance of disambiguation.
To evaluate disambiguation accuracy, we masked 5\% of the entries of the adjacency matrix $A$ before decomposing it (\cf\ \Secref{sec:Link densification}).
Each masked link is associated with a potentially ambiguous anchor text~$p$.
Given $p$, we generated all candidate targets $j$ and ranked them by their score $u_i v_j^\top$.
Disambiguation accuracy is then defined as the fraction of masked matrix entries for which the top-ranked candidate was correct.
It is summarized, for all 28 languages, in the 4 rightmost columns of \Tabref{tab:langs}, where column ``$[l,u]$'' contains the accuracy for anchors with at least $l$ and at most $u$ candidates.

The column ``$[1,\infty]$'' shows the overall accuracy for all anchors (85\% on average over all 28 languages).
Since this column includes trivial, unambiguous anchors, the column ``$[2,\infty]$'' is more interesting.
Although lower, these numbers are still satisfactorily high (69\% on average), particularly when compared to the random baseline~(24\%).

Manual error analysis revealed that anchors with a large number of candidates tend to be inherently hard to disambiguate even for humans (\eg, ``self\hyp titled album'' has 712 candidates).
Hence, preferring precision over recall,
our implementation ignores
phrases with more than 10 candidates,
obtaining an
average accuracy of 87\% for the remaining anchors (``$[1,10]$'').

Moreover, we found that, even when the exact link target was not identified, often a semantically close target was predicted. Since this suffices for our purpose---which is not to disambiguate all links perfectly, but rather to create better document representations for the downstream topic modeling step---the accuracy of 87\% should be considered an underestimate of the true, effective utility.

The quality of disambiguated links is confirmed by the superior performance of densified, compared to raw, bags of links, as discussed next.


\subsection{Topic modeling}
\label{sec:evaluation: topic modeling}

We evaluated 4 model classes, each trained on a different corpus:
\begin{enumerate}
\item \textbf{WikiPDA, dense links, 28 languages:} full model as described in \Secref{sec:WikiPDA}.
\item \textbf{WikiPDA, sparse links, 28 languages:} the same, but without link densification.
\item \textbf{WikiPDA, dense links, English:} trained on English only, rather than on all 28 languages.
\item \textbf{Text-based LDA, English:} bag-of-words LDA trained on English text (not links).
\end{enumerate}
For each model class, we trained and evaluated models for 10 values of $K$, ranging from 20 to 200.
In the following, ``model'' refers to an instance of a model class trained for a specific $K$.

\begin{figure}
\includegraphics[width=0.47\textwidth]{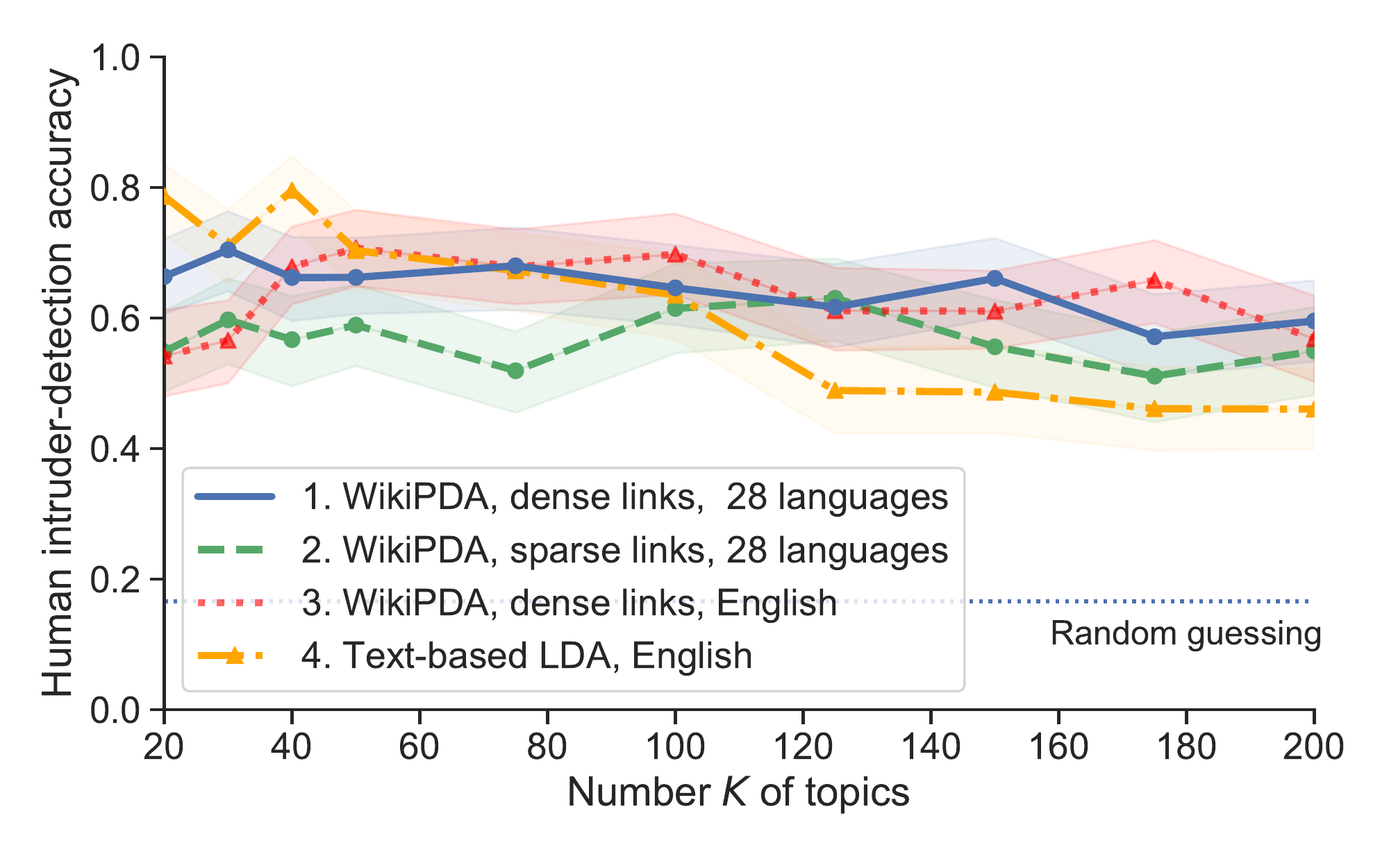}
\vspace{-4mm}
\caption{
Evaluation of topic models.
Topic coherence measured in terms of human intruder\hyp detection accuracy (higher is better), with 95\% confidence intervals.
}
\label{fig:human_accuracy}
\end{figure}

Comparing model classes 1 and 2 lets us determine the benefits of link densification; comparing model classes 1 and 3, whether including more languages hurts performance, as has been found to be the case in other \cl settings \cite{josifoski2019crosslingual}; and comparing model classes 3 and 4, whether using bags of links rather than bags of words makes a difference.

\xhdr{Methodology: intruder detection}
The evaluation of topic models is challenging.
Traditionally, it has been based on automatic scores such as perplexity, capturing how ``surprising'' documents from a held-out corpus are, given the training set.
Unfortunately, perplexity does not necessarily correlate with human judgment, and in some cases an inverse relation has even been reported \cite{NIPS2009_3700}. 
Since we are interested in interpretable models, we quantified the utility of topics in a human, rather than automatic, evaluation,
using the \textit{word intruder} framework proposed by \citet{NIPS2009_3700}.
Given a model to evaluate, we randomly selected $n=20$ of the $K \geq 20$ topics and extracted the top 5 Wikidata concepts per topic.
Then we selected an \textit{intruder concept} for each topic:
a concept that ranked low for that topic, but high for at least one other topic
(in particular, the concept with the largest rank difference was selected).
A shuffled list of the 6 concepts (described by their English names) was shown to a human evaluator, who was asked to spot the intruder.
The more coherent a topic, the easier it is to spot the intruder, so human accuracy serves as a measure of topical coherence.


\xhdr{Crowdsourcing setup}
For each model, human accuracy was estimated based on $12n=240$ workers' guesses obtained from 12 independent rounds of the above procedure on Amazon Mechanical Turk.
Workers were shown a page with instructions and a batch of 17 intruder\hyp detection tasks: 16 regular tasks (4 per model class), each with a different $K$, and one control task with an obvious answer to assess worker reliability (all workers were found to be reliable). Workers were encouraged to search online in case they did not know the meaning of a concept.
To not reveal a pattern, we used each model and each intruder at most once per batch.

\xhdr{Results}
\Figref{fig:human_accuracy} shows that,
with the full WikiPDA model (model class 1), human intruder\hyp detection accuracy was 60--70\%, depending on $K$, far above random guessing (16.7\%).

Comparing model classes 1 and 2, we see that the dense WikiPDA model yielded results consistently above the sparse model (by up to 15 percentage points), showing the utility of link densification.

Comparing model classes 1 and 3, we find that the dense WikiPDA model for 28 languages performed indistinguishably from the dense model for English only; \ie, adding more languages did not make the topics less coherent.
This outcome is noteworthy, since on other \cl tasks (\eg, document retrieval), performance on a fixed testing language decreased when adding languages to the training set \cite{josifoski2019crosslingual}.

Comparing model classes 3 and 4 (both English only) shows that, whereas the performance of text-based LDA degrades with growing $K$, WikiPDA is more stable.
While text-based LDA is slightly better for small $K \leq 50$, WikiPDA prevails for $K \geq 75$.
This suggests that the link-based models are more customizable to use cases where the problem requires a specific $K$.

Note that the text-based LDA model is not language\hyp independent and thus not truly a competitor with crosslinguality in mind.
Rather, it should \textit{a priori} be considered a strong ceiling:
text-based LDA is the de-facto standard for analyzing the content of \WP articles in monolingual settings \cite{Singer2017-dt,lemmerich2019world}.
Thus, by surpassing the topical coherence of text-based models, WikiPDA offers crosslinguality ``for free''.


\section{Applications}
\label{sec:Applications}

WikiPDA enables a wide range of applications,
some of which we spotlight next.
We emphasize that the purpose of this section is not to take a deep dive into specific directions, but rather to exemplify the utility of WikiPDA as a general tool for analyzing \WP across languages.

\begin{figure}
\includegraphics[width=0.47\textwidth]{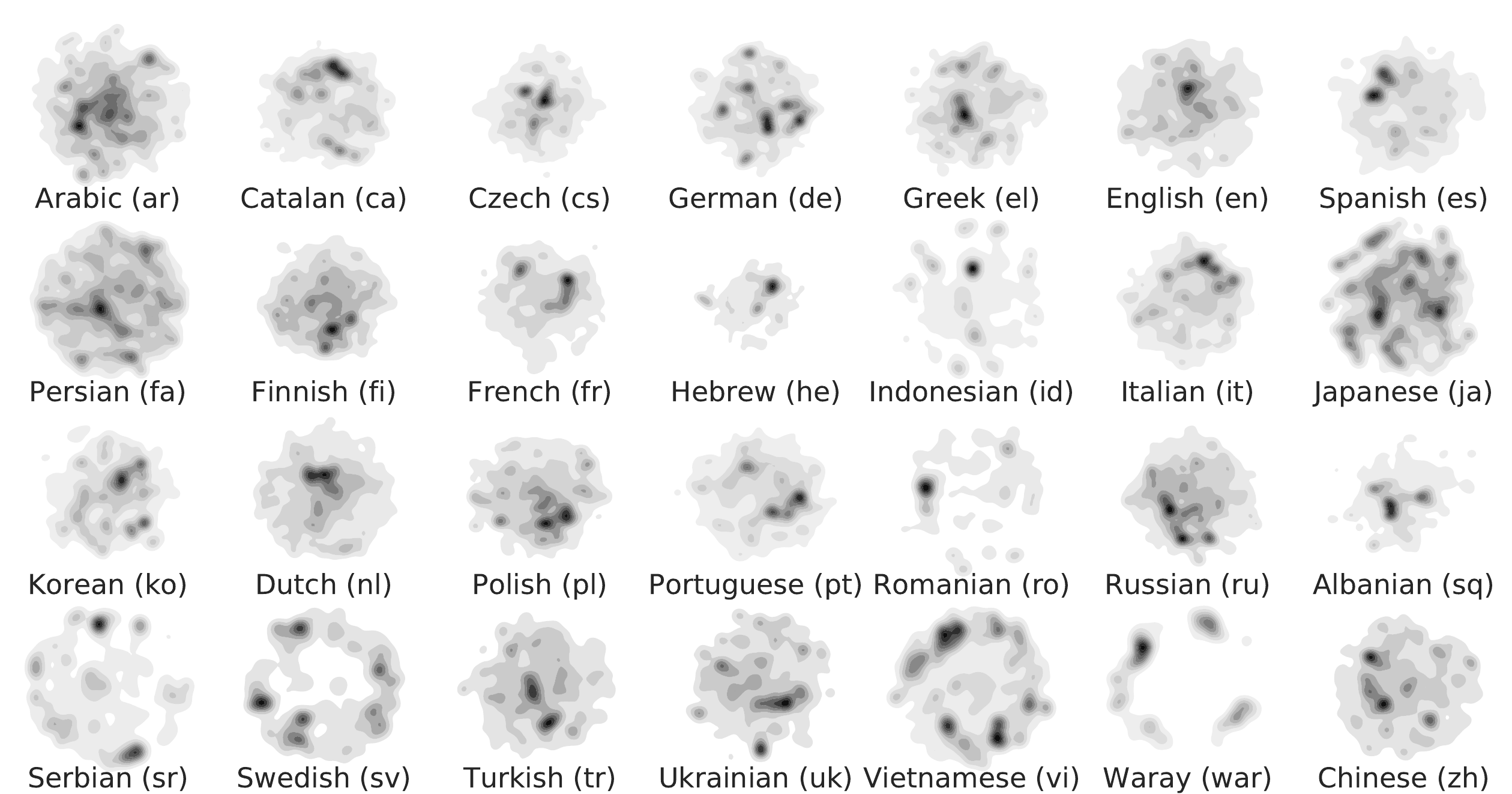}
\caption{
Heat-map visualization of the topic distributions of 28 \WP language editions, obtained by reducing the dimensionality of the topic vectors from $K=40$ down to 2 dimensions via t-SNE \cite{maaten2008visualizing}.
The visual heterogeneity of the heat maps highlights the topical heterogeneity of the various language editions.
}
\label{fig:2d_projections}
\end{figure}

\subsection{Comparing Wikipedia across languages}
\label{sec:Comparing Wikipedia across languages}

\WP's different language editions are maintained by independent volunteer communities, each with their own cultural background and with potentially diverging interests.
Understanding the differences in content coverage across  \WP language editions constitutes a major topic for researchers in multiple domains \cite{bao2012omnipedia,callahan2011cultural,hecht2009measuring,hecht2010tower,laufer2015mining,massa2012manypedia,roy2020topic},
and WikiPDA will be a useful tool for their endeavors.

\xhdr{Topical bias}
Using WikiPDA, we studied the topical bias of 28 language editions (\cf\ \Tabref{tab:langs}).
To obtain a first impression, we pooled the topic vectors from all languages and reduced their dimensionality from $K=40$ down to 2 dimensions via t-SNE \cite{maaten2008visualizing}.
A heat-map visualization of the reduced, 2-dimensional topic vectors for each language is presented in \Figref{fig:2d_projections}.
The visual heterogeneity of the heat maps is a stark indication of the topical heterogeneity of the various language editions.

\begin{figure*}
  \includegraphics[width=\textwidth]{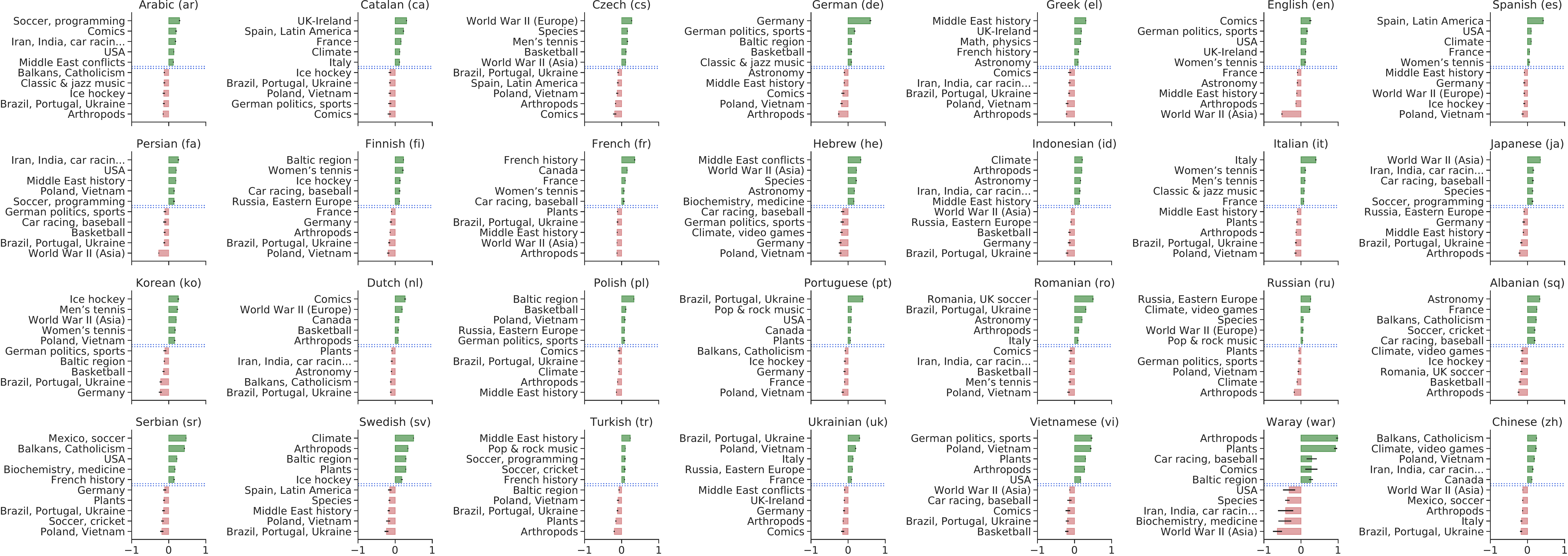}
  \caption{
  Topical bias of 28 \WP language editions.
  For each language $L$, a logistic regression was trained to predict if an article was written in language $L$, using the article's distribution over WikiPDA topics (labeled manually with names) as predictors.
  Most predictive positive and negative coefficients are shown, with 95\% confidence intervals.
  }
  \label{fig:langpred_top_features}
\end{figure*}

Whereas reducing the dimension from 40 to 2 is advantageous for visual inspection, the 2 dimensions resulting from t-SNE---unlike the $K=40$ original dimensions---do not have a clear interpretation anymore.
In order to study differences across languages with respect to individual topics, we therefore conducted a regression analysis.
For each language $L$, we randomly sampled 20K articles as positive examples and
$20\text{K}/27=740$
from each of the 27 other languages as negative examples, applied an 80/20 train\slash test split, and trained a one-\vs-all logistic regression classifier to predict whether an article is from language $L$, given the article's topic distribution.%
\footnote{Since in logistic regression the log odds are modeled as a linear function of the predictors, we also expressed the probabilistic predictors (namely, the $K$ topic probabilities) as log odds, so input and output use the same ``units''.}
On average the 28 classifiers achieved an
area under the ROC curve (AUC)
of 78\% for $K=40$, or 84\% for $K=200$, significantly above the random baseline of 50\%, indicating major differences across language editions.
Inspecting the fitted coefficients for $K=40$, depicted in \Figref{fig:langpred_top_features}, revealed the specificities of individual languages.
First and foremost, country- or region\hyp specific topics appeared among the most discriminative topics.
Additionally, several more surprising associations emerged:
\eg, \cpt{Comics} is the topic most indicative of English and Dutch, and it is most counter\hyp indicative of Ukrainian and Catalan;
\cpt{Geopolitics} is prominently featured in Hebrew;
\cpt{Ice hockey} and \cpt{Tennis}, in Korean; \etc

Note that some topics appear to conflate multiple concepts that one would rather expect to emerge as distinct topics of their own (\eg, the topic \cpt{Poland, Vietnam}).
Such conflation is expected for the small number of $K=40$ topics used to produce \Figref{fig:langpred_top_features}.
Although such a small $K$ is convenient as it allows for the manual inspection and naming of topics, it provides too little capacity to capture all of Wikipedia's diversity.
Accordingly, we found that the conflation effect is reduced as $K$ grows.
Also note that conflation does not lead to catch-all ``garbage'' topics, but to topics consisting of distinct subtopics (\eg, \cpt{Poland} and \cpt{Vietnam}). The fact that subtopics are dissimilar is in fact desirable: we found that, when allowing for a larger model capacity (larger $K$), they tend to become distinct, non-redundant topics of their own.

\begin{figure}[tb]
\centering
\subfloat[All articles]{\includegraphics[width=.8\columnwidth]{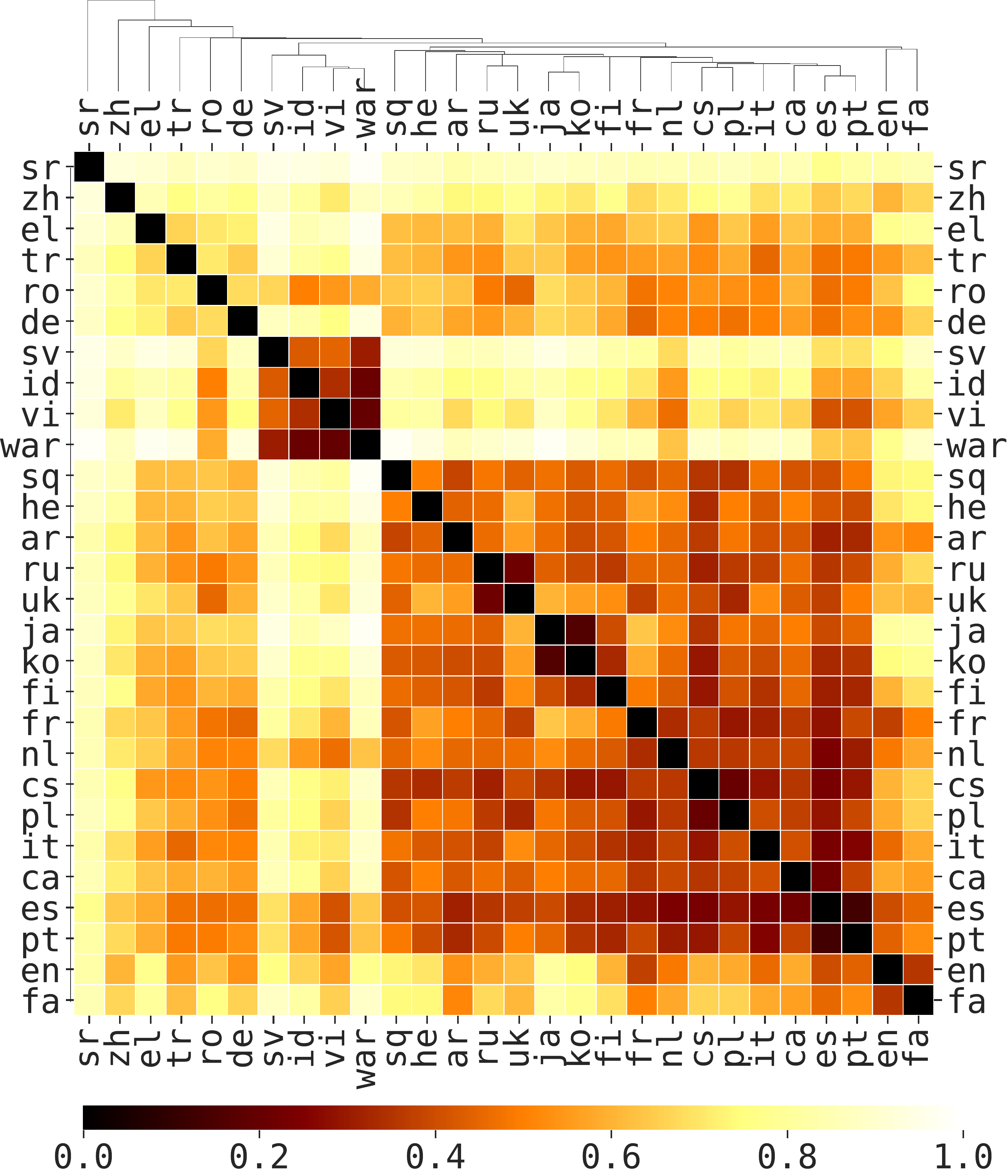}\label{fig:distance_all_articles}}\\
\subfloat[Common articles]{\includegraphics[width=.8\columnwidth]{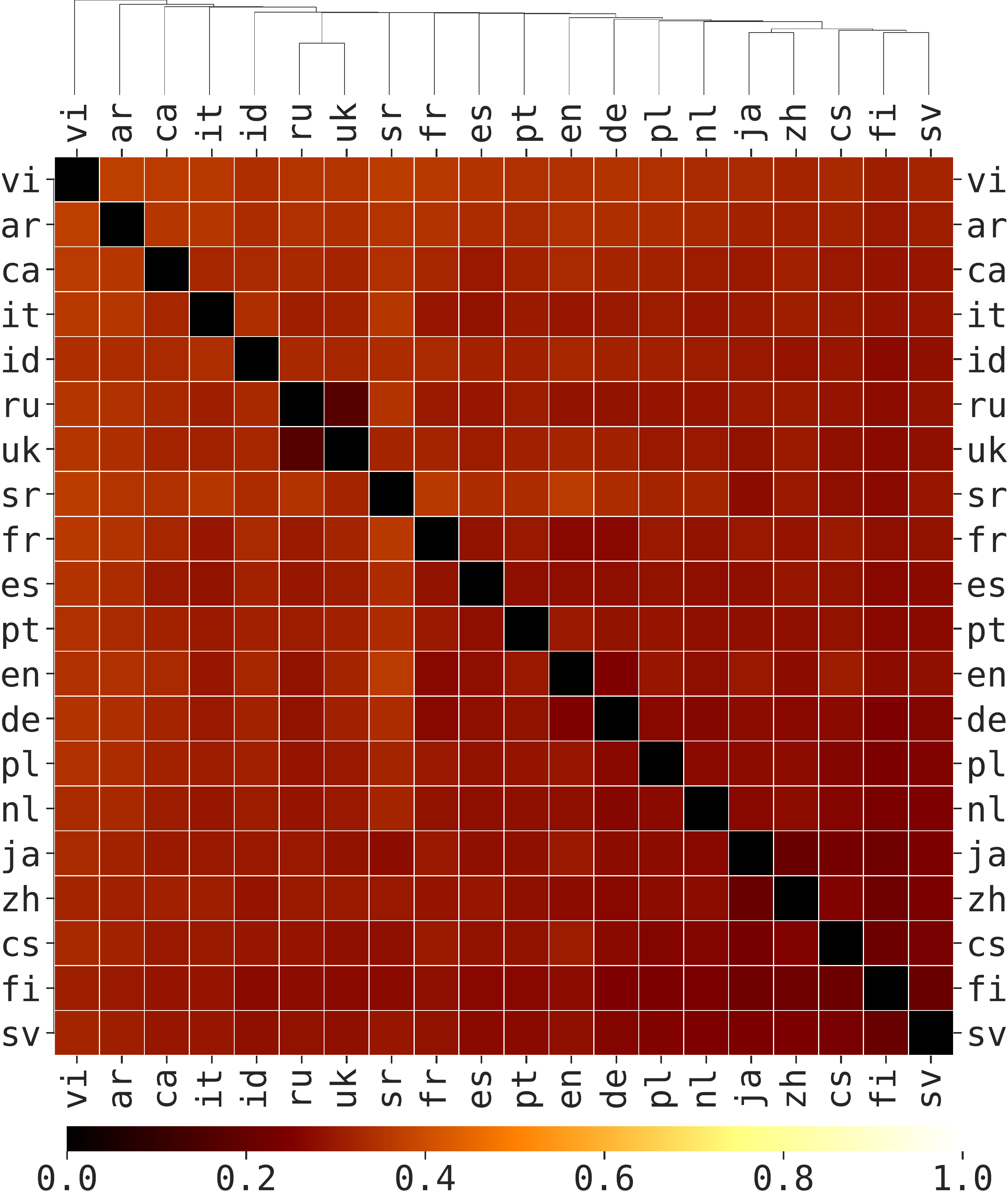}\label{fig:distance_intersection}}
\vspace{-3mm}
\caption{
Cosine distance between \WP language editions.
\textbf{(a)} 28 languages, each represented via average topic vector of all articles.
\textbf{(b)} 20 top languages, considering only the 16K articles included in all 20 languages.
    }
    \label{fig:language_distance}
\end{figure}

\xhdr{Distance between language editions}
Next, we computed pairwise distances for all language editions via the cosine distance between the languages' mean topic vectors.
As we do not rely on topic labels here, we use the larger $K=200$.
\Figref{fig:distance_all_articles} shows the distance matrix.
Rows and columns correspond to languages, sorted by performing agglomerative clustering based on the distances and listing the leaves of the dendrogram in left-to-right order, such that the most similar pairs cluster along the diagonal of the distance matrix.
Clear topical similarities (darker colors) emerge for languages of countries with historical or geographical ties, including Japanese\slash Korean, Russian\slash Ukrainian,
Czech\slash Polish, and Portuguese\slash Spanish.
Waray (spoken in the Philippines) clusters with Indonesian, Vietnamese, and---more surprisingly---Swedish, a language that, linguistically speaking, could not be more distant.
Investigating the reasons, we found that Swedish and Waray are among the three Wikipedias (the third being Cebuano) in which Lsjbot was active, a bot that created 80--99\% of the articles in those languages.
\Figref{fig:langpred_top_features} suggests that Lsjbot created particularly many biology\hyp related articles (with \cpt{Arthropods} and \cpt{Plants} appearing as prominent topics in both Swedish and Waray), a finding not even mentioned on the \WP page about Lsjbot itself.%
\footnote{\url{https://en.wikipedia.org/w/index.php?title=Lsjbot&oldid=949492392}}
Also, it seems that the bot, which was created by a Swede, gave Waray \WP a Swedish bias, as indicated by Waray's large coefficient for the topic \cpt{Baltic region} in \Figref{fig:langpred_top_features}.
These nuggets exemplify how WikiPDA enables the cross\hyp cultural study of \WP.

The above\hyp noted differences may be due to different language editions containing articles about different concepts.
An equally interesting question asks to what extent the languages differ in how they discuss identical concepts.
To quantify this, we found the 16K articles in the intersection of the 20 largest language editions and computed, for each language pair
and each common article,
the cosine distance of the two languages' topic vectors for the article.
Averaging the 16K distances yields \Figref{fig:distance_intersection}, which paints a more uniform picture than \Figref{fig:distance_all_articles},
with no important differences remaining between languages.
Note, however, that
Russian\slash Ukrainian,
Finnish\slash Swedish,
and Chinese\slash Japanese
cover the same concepts in particularly similar ways.


\begin{figure*}[htbp]
\centering
\subfloat[28 languages seen during WikiPDA training]{\includegraphics[height=3.4cm]{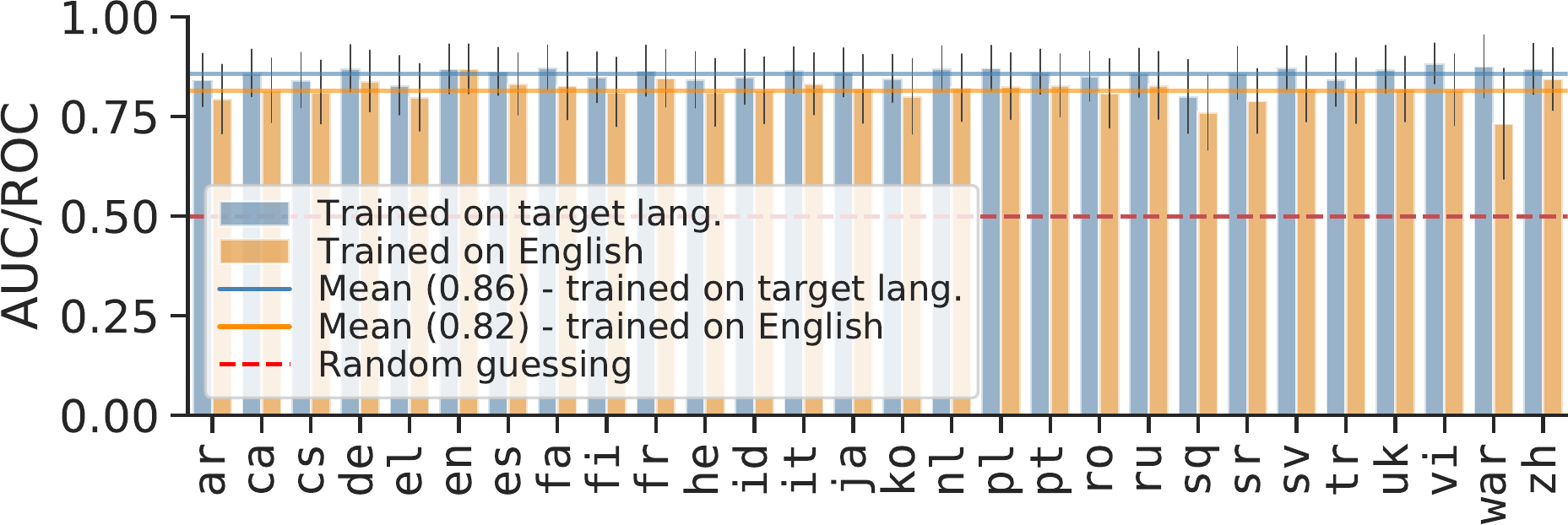}\label{fig:supervised performance 28 lang}}
\hspace*{0.5cm}
\subfloat[17 zero-shot languages]{\includegraphics[height=3.4cm]{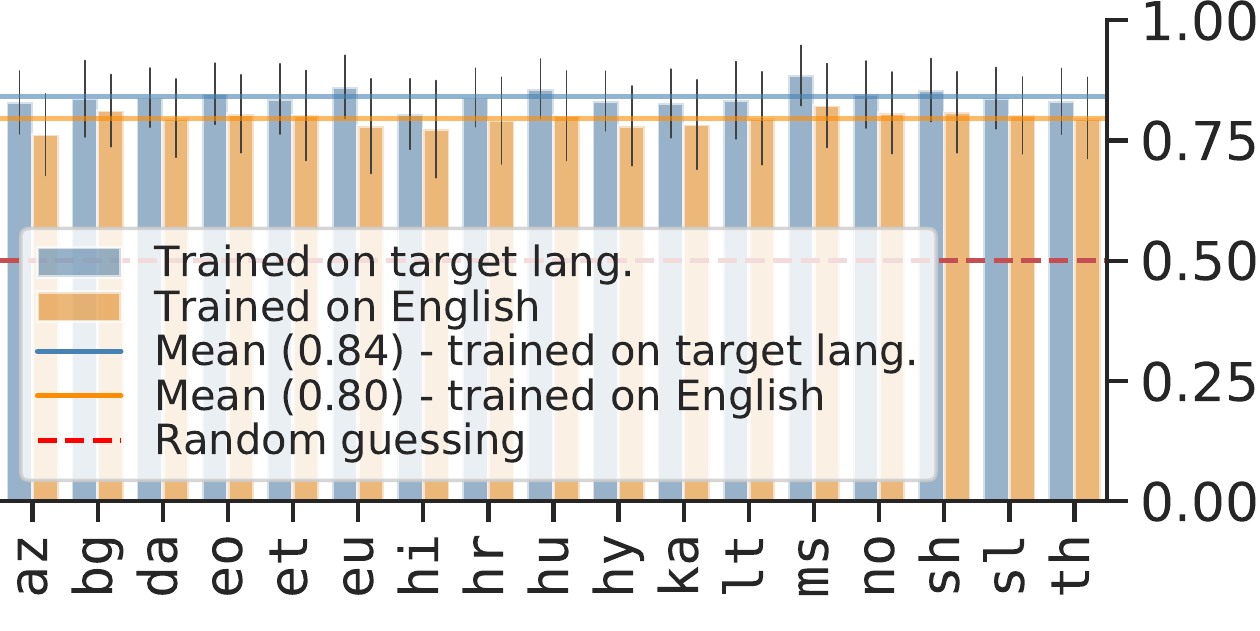}\label{fig:supervised performance 17 lang}}
\vspace{-3mm}
\caption{Performance on supervised topic classification, using unsupervised WikiPDA topics as features.
    For each language~$L$, two models were evaluated:
    trained on $L$ (blue);
    trained on English (orange).
    Error bars: standard deviation over 64 binary classification tasks (one per supervised topic label).
    Similarity of blue and orange shows that classifier works on languages not seen during supervised classifier training.
    Similarity between (a) and (b) shows that classifier and WikiPDA models work on languages not seen during unsupervised WikiPDA training.}
\end{figure*}

\subsection{Supervised topic classification}
\label{sec:Supervised topic classification}


WikiPDA is an unsupervised technique.
It discovers whatever topics are best suited for summarizing the data.
Sometimes, however, researchers may want to exert more control by fixing a set of topics ahead of time and classifying documents into those in a supervised manner.
For instance, with the ORES library,%
\footnote{\url{https://ores.wikimedia.org}}
Wikimedia provides a supervised classifier for categorizing English articles into a manually constructed taxonomy of 64 topics, based on features derived from the articles' English text \cite{asthana2018few}.
We explored if WikiPDA topic vectors can be used as features instead, giving rise to a language\hyp independent model, whereas the ORES model is language\hyp specific.
More generally, we establish whether WikiPDA is effective as an unsupervised pre\hyp training step for supervised downstream tasks.

\xhdr{Setup}
For training and testing (following an 80/20 split), we used
the same dataset that had been used to train the ORES topic model,%
\footnote{Code: \url{https://github.com/wikimedia/drafttopic}}
consisting of 5.1M English articles, each labeled with 64 binary labels that specify for each of the 64 topical classes defined by ORES whether the article belongs to the class.
The labels were obtained by the creators of ORES by manually mapping WikiProjects%
\footnote{\url{https://en.wikipedia.org/wiki/Wikipedia:WikiProject}}%
---and thus implicitly their constituent articles---to the 64 high-level topics.
Note that, although the ORES training data consists of English articles only, Wikidata's crosslingual alignment allows us to propagate labels to other languages for any article that has an English version. For this practical reason, our evaluation focuses on such articles.

Each article can belong to multiple classes, so we trained an independent binary logistic regression classifier per class, on a balanced training set where negative examples were sampled evenly from the 63 other classes.
Performance was found to increase with $K$, so we used $K=200$.
For each language~$L$, we performed two evaluations:
first, with a model trained on articles from $L$ (after transferring labels from the English dataset via the alignment given by Wikidata)
and second, with a model trained on English.


\xhdr{Results}
In \Figref{fig:supervised performance 28 lang}, we show two AUC values (macro\hyp averages over the 64 classes) for each language $L$: one when testing the classifier trained on $L$ itself (blue); the other, when testing the classifier trained on English (orange).
Performance is high across all languages, with an average AUC of 86\% for the language\hyp specific classifiers.
The single, fixed classifier trained on English performed only slightly worse when evaluated on the other languages, with an average AUC of 82\%.

In English, the easiest class was \cpt{Video games} (AUC 97\%); the hardest class was \cpt{Biographies} (AUC 74\%).

Note that the primary goal of these experiments was not to achieve maximum classification performance by all means.
Indeed, exploratory results showed that simply switching from logistic regression to gradient\hyp boosted trees immediately boosted the AUC by 2--3 percentage points,
and we would expect methods that leverage state-of-the-art pre-trained language models such as BERT \cite{devlin2018bert} to perform even better. We emphasize that our main goal is to discover interpretable topics in an unsupervised fashion and that the supervised application presented here is primarily intended as an additional evaluation to assess the usefulness of the learned representations, not so much as an end goal in itself.

In this light, the main take-aways of this section are twofold:
(1)~WikiPDA's unsupervised topics can be readily translated to a different set of manually defined topics, which demonstrates their utility as a general low\hyp dimensional representation that captures the topical essence of a document.
(2)~Due to the \cl nature of WikiPDA topics, a supervised model trained on one language (here: English) can be transferred to any other language not seen during supervised training, achieving high performance without any fine\hyp tuning.

In our final set of experiments, described in the next section, we push the language\hyp transfer paradigm even further, by moving to a setting where the target language was absent not only during training of the supervised classifier, but also during unsupervised training of the WikiPDA topics that the supervised classifier uses as features.



\section{Zero-shot language transfer}
\label{sec:Zero-shot language transfer}

The bags of links by which WikiPDA represents input documents are composed of language\hyp independent Wikidata concepts (one per out-link).
No matter in what language an article is written, its bag of links can be immediately compared to the bags of links extracted from any other language.
This way, a WikiPDA model trained on a certain set of languages can be used to infer topics for articles from any new language not seen during WikiPDA training.
In other words, WikiPDA inherently enables \textit{zero-shot language transfer.}
This capability is particularly convenient for low\hyp resource languages, where the available data might not suffice to learn meaningful topics,
and it sets WikiPDA apart from all the previously proposed \cl topic models discussed in \Secref{sec:Related work}, which need to be retrained whenever a new language is added.

To showcase WikiPDA's zero-shot capability, we used the model trained on the 28 languages of \Tabref{tab:langs} to infer topics for all articles in 17 more languages
(\cf\ \Figref{fig:supervised performance 17 lang})
and repeated the supervised topic classification experiments
(\Secref{sec:Supervised topic classification}) for these languages.
As in \Secref{sec:Supervised topic classification}, we evaluated two supervised topic classifiers for each language:
one trained on the respective language,
the other trained on English.
Note that in neither case were the 17 new languages included in topic model training;
rather, the topic vectors that served as input to the supervised classifier were inferred based on a WikiPDA model trained only on the 28 old languages.
Despite this, the mean AUC on the 17 new languages (\Figref{fig:supervised performance 17 lang}) nearly reached that on the 28 old ones, both for the language\hyp specific classifiers (84\% for the new languages, \vs\ 86\% for the old languages)
and for the English classifier (80\% \vs\ 82\%).

Finally, to further validate the applicability of WikiPDA topics in the zero-shot setting, we repeated the analysis of \Figref{fig:langpred_top_features},
fitting logistic regression models to predict the language of an article given its topic vector.
Classification performance was as high on the 17 new languages as on the 28 languages seen during topic model training (mean AUC 79\% for $K=40$; 84\% for $K=200$), indicating that the topic vectors capture the peculiarities of the 17 new languages well, even though the languages were not seen during topic model training.



\section{Discussion and conclusion}
\label{sec:Discussion}

We presented Wikipedia\hyp based Polyglot Dirichlet Allocation (WikiPDA), a novel \cl topic model for \WP,
founded on the fact that \WP articles are laced with language\hyp independent links to other articles.
Our human evaluation showed that the topics learned from 28 languages are as coherent as those learned from English alone, and more coherent than those from text-based LDA on English, a noteworthy finding, given that other \cl tasks have suffered by adding languages to the training set \cite{josifoski2019crosslingual}.
We demonstrated WikiPDA's practical utility in two example applications and highlighted its capability for zero-shot language transfer.

The key insight underpinning WikiPDA is that, when taken as bags of links, \WP articles are \cl from the get-go, leading to two big advantages, interpretability and scalability:

\xhdr{Interpretability}
By invoking a probabilistic topic model such as LDA as a subroutine, WikiPDA inherits all advantages of that model, including the interpretability of topics as distributions over terms (in our case, Wikidata concepts) and of documents as distributions over topics.
Even more important, as WikiPDA's vocabulary consists of Wikidata concepts, which have names in all languages, bags of links and learned topics (distributions over the vocabulary), can be interpreted even without understanding the corpus languages.
WikiPDA therefore confers distinct advantages, compared to black-box models such as those trained via deep learning, including recurrent neural networks and transformers.
Although such models might perform better at specific downstream tasks, they could not be considered valid alternatives if the user expects to obtain interpretable document representations.
On the contrary, it is precisely such users whom WikiPDA is intended to serve.

\vfill
\xhdr{Scalability}
In bag-of-links space, the corpus can be treated as monolingual, such that standard topic models apply, for which highly efficient algorithms exist;
\eg, online algorithms for LDA can handle massive amounts of text \cite{hoffman2010online} and have been implemented for high\hyp performance machine learning libraries (\eg, Vowpal Wabbit)
and massively parallel big data platforms (\eg, Apache Spark).
Scaling WikiPDA to all of Wikipedia's 299 languages is thus fully within reach, whereas previous methods
(\Secref{sec:Related work})
have usually been deployed on small sets of languages only.
That said, given WikiPDA's zero-shot capability (\Secref{sec:Zero-shot language transfer}), training on all 299 languages may not even be necessary, as a model trained on a few languages can be immediately applied to unseen languages ``for free''.

\vfill
\xhdr{On the use of machine translation}
Instead of making documents language\hyp agnostic by mapping them to bags of links, one might be tempted to pursue an alternative approach where all documents would first be machine\hyp translated to a pivot language.
While feasible in principle, we consider such an approach to run counter to our design principles.
First, translating entire \WP editions would be a costly endeavor that many researchers cannot afford (indeed, much of the literature on crosslingual document representations starts from the very desideratum to circumvent machine translation), whereas extracting and denisfying links from all \WP articles is efficiently feasible.
Second, the Wikipedia editions that could benefit most from \cl topic models in order to discover knowledge gaps and content skewness are the low-resource languages, for many of which no machine translation models exist to begin with; \eg, we are not aware of translation models for the Waray language, although it has the 11th-largest Wikipedia edition and features prominently in our analysis.

\vfill
\xhdr{Limitations}
Finally, we discuss two potential concerns: language imbalance and link sparsity.

First, \WP's language editions vary considerably in size, so the learned topics are dominated by larger languages.
Whether this is desirable or not depends on the specific use case.
Future work should explore the effects of upweighting smaller languages,
\eg, by downsampling large languages, upsampling small languages, or incorporating weights into LDA's objective function.
Alternatively, one could aggregate articles at the concept level, \eg, by unioning the bags of links of the articles about the same Wikidata concept in different languages.

Second, compared to text-based models, our link-based model works with sparser inputs, even after link densification.
While advantageous computationally, this raises two questions:
(1)~whether the set of Wikidata concepts is rich enough to capture all semantic facets of an input document;
and (2)~whether WikiPDA can handle very ``short'' documents, \ie, articles with very few outgoing links.
Regarding the first question, we emphasize that, with over 32M entities, Wikidata is large enough to cover most vocabulary entries.
In this light, moving from words to Wikidata entities may be seen as additionally offering some common NLP preprocessing steps for free: the removal of stop words and rare $n$-grams, plus lemmatization.
Regarding the second question, we trained and tested the supervised topic classification model (\Secref{sec:Supervised topic classification}) for English again, this time only on articles with fewer than 10 links (19\% of all articles).
The model still performed well (AUC 85\%),
only 2 percentage points lower than when using all articles, including those with many links,
indicating that WikiPDA is not hampered in important ways by articles with few links.

\xhdr{Going beyond \WP}
In the version presented here, WikiPDA is specifically geared toward the analysis of \WP articles as input documents, since the matrix\hyp completion\hyp based link densification method requires at least a few pre\hyp existing \WP links in each input document.
Pushing further, it will be interesting to extend WikiPDA to work on documents without any \WP links whatsoever.
We believe this is well within reach, by exploiting WikiPDA's modularity:
any method that annotates documents with links to a knowledge base can be plugged in instead of matrix completion, including sophisticated entity linking methods for linking mentions in plain-text documents (\ie, without pre-existing links to Wikidata) to a knowledge base  \cite{kolitsas2018end,shen2014entity}.
This would widen the scope of applicability to many document types, as long as the documents contain entity mentions that can be linked to the knowledge base.
Moreover, past work \cite{West2010-nl} has shown that plain-text-based entity linkers can be fruitfully combined with matrix completion as used in this paper (as postprocessing), and we would expect similar outcomes for WikiPDA.

Such extensions would allow us to answer many important questions outside of the realm of \WP.
For instance, given a corpus of school textbooks in many languages, we may ask: how do different countries' curricula treat the same subjects (e.g., World War~II)?
Given a corpus of COVID-related news from all over the world, we may ask: on what aspects of the pandemic have news outlets in different countries focused over time?
Given a corpus of search engine results for queries about climate change in different languages, we may ask: is there a biased view on the issue based on user language?
In all of these examples, we require a common, interpretable set of topics across languages, which monolingual topic models cannot provide. 
As a way forward, off-the-shelf entity linkers can be used as preprocessing to annotate the corpus with links to Wikipedia articles; WikiPDA can take over from there. 
During training, topics can be learned either on a \WP corpus, as proposed in this paper, or on the corpus of downstream study itself, if preferred.

That said, as laid out in the introduction, Wikipedia on its own constitutes such an important use case that, even without expanding the scope of WikiPDA beyond Wikipedia articles, it still provides a highly valuable tool for a large research community.

\xhdr{Conclusion}
To conclude,
WikiPDA offers researchers a practical tool for studying the content of all of \WP's 299 language editions in a unified framework, thus better reflecting \WP's real diversity.
We look forward to seeing it deployed in practice.

\section*{Acknowledgments}
{\small
We thank Ahmad Ajalloeian and Blagoj Mitrevski for their valuable early contributions, and Daniel Berg Thomsen for help with library development.
This project was partly funded by
the Swiss National Science Foundation (grant 200021\_185043),
the European Union (TAILOR, grant 952215),
and the Microsoft Swiss Joint Research Center.
We also gratefully acknowledge generous gifts from Facebook and Google.
}

\balance
\bibliographystyle{ACM-Reference-Format}
\bibliography{main}


\end{document}

%% file: dlab_macros.tex
\usepackage[utf8]{inputenc}
\usepackage[T1]{fontenc}
\usepackage{hyphenat}
\usepackage{xspace}
\usepackage{amsmath}
\usepackage{amsfonts}
\usepackage{hyperref}
\usepackage{url}
\usepackage{booktabs}
\usepackage{multirow}
\usepackage{subfig}
\usepackage{makecell}
\usepackage{caption}
\usepackage{minibox}
\usepackage{bbm}
\usepackage{graphicx}
\usepackage{balance}
\usepackage{mathtools}
\usepackage{color}
\usepackage{marvosym}
\usepackage{ifthen}
\usepackage{textcomp}
\usepackage{enumitem}
\usepackage{verbatim}
\usepackage{algorithm}
\usepackage{algorithmic}
\usepackage{numprint}
\usepackage{balance}

\usepackage{amsthm}
\theoremstyle{plain}

\newcommand{\chatoDisplayMode}[1]{#1}

\definecolor{MyRed}{rgb}{0.6,0.0,0.0} 
\definecolor{MyBlack}{rgb}{0.1,0.1,0.1} 
\newcommand{\inred}[1]{{\color{MyRed}\sf\textbf{\textsc{#1}}}}
\newcommand{\frameit}[2]{
  \begin{center}
  {\color{MyRed}
  \framebox[.9\columnwidth][l]{
    \begin{minipage}{.85\columnwidth}
    \inred{#1}: {\sf\color{MyBlack}#2}
    \end{minipage}
  }\\
  }
  \end{center}
}

\newcommand{\note}[2][]{\chatoDisplayMode{\def\@tmpsig{#1}\frameit{{\Pointinghand} Note}{#2\ifx \@tmpsig \@empty \else \mbox{ --\em #1}\fi}}}
\newcommand{\todo}[2][]{\chatoDisplayMode{\def\@tmpsig{#1}\frameit{{\Writinghand} To-do}{#2\ifx \@tmpsig \@empty \else \mbox{ --\em #1}\fi}}}

\newcommand{\abbrevStyle}[1]{#1}

\newcommand{\ie}{\abbrevStyle{i.e.}\xspace}
\newcommand{\eg}{\abbrevStyle{e.g.}\xspace}
\newcommand{\cf}{\abbrevStyle{cf.}\xspace}

\newcommand{\vs}{\abbrevStyle{vs.}\xspace}
\newcommand{\etc}{\abbrevStyle{etc.}\xspace}

\newcommand{\Secref}[1]{Sec.~\ref{#1}}

\newcommand{\Tabref}[1]{Table~\ref{#1}}
\newcommand{\Figref}[1]{Fig.~\ref{#1}}

\newcommand{\xhdr}[1]{\vspace{1.7mm}\noindent{{\bf #1.}}}

\newcommand{\xhdrNoPeriod}[1]{\vspace{1.7mm}\noindent{{\bf #1}}}

\newcommand{\textcite}[1]{\citeauthor{#1} \shortcite{#1}}

\newcommand{\cpt}[1]{\textsc{\MakeLowercase{#1}}}

\newcommand{\hide}[1]{}

\DeclareMathOperator*{\argmax}{arg\,max}

\makeatletter
\newcommand{\iffont}[2]{\ifthenelse{\equal{\f@family}{#1}}{#2}{}}
\makeatother

  \DeclareSymbolFont{greek}{OML}{cmm}{m}{n}
  \DeclareMathSymbol{\alpha}{\mathalpha}{greek}{"0B}
  \DeclareMathSymbol{\beta}{\mathalpha}{greek}{"0C}
  \DeclareMathSymbol{\gamma}{\mathalpha}{greek}{"0D}
  \DeclareMathSymbol{\delta}{\mathalpha}{greek}{"0E}
  \DeclareMathSymbol{\epsilon}{\mathalpha}{greek}{"0F}
  \DeclareMathSymbol{\zeta}{\mathalpha}{greek}{"10}
  \DeclareMathSymbol{\eta}{\mathalpha}{greek}{"11}
  \DeclareMathSymbol{\theta}{\mathalpha}{greek}{"12}
  \DeclareMathSymbol{\iota}{\mathalpha}{greek}{"13}
  \DeclareMathSymbol{\kappa}{\mathalpha}{greek}{"14}
  \DeclareMathSymbol{\lambda}{\mathalpha}{greek}{"15}
  \DeclareMathSymbol{\mu}{\mathalpha}{greek}{"16}
  \DeclareMathSymbol{\nu}{\mathalpha}{greek}{"17}
  \DeclareMathSymbol{\xi}{\mathalpha}{greek}{"18}
  \DeclareMathSymbol{\pi}{\mathalpha}{greek}{"19}
  \DeclareMathSymbol{\rho}{\mathalpha}{greek}{"1A}
  \DeclareMathSymbol{\sigma}{\mathalpha}{greek}{"1B}
  \DeclareMathSymbol{\tau}{\mathalpha}{greek}{"1C}
  \DeclareMathSymbol{\upsilon}{\mathalpha}{greek}{"1D}
  \DeclareMathSymbol{\phi}{\mathalpha}{greek}{"1E}
  \DeclareMathSymbol{\chi}{\mathalpha}{greek}{"1F}
  \DeclareMathSymbol{\psi}{\mathalpha}{greek}{"20}
  \DeclareMathSymbol{\omega}{\mathalpha}{greek}{"21}
  \DeclareMathSymbol{\varepsilon}{\mathalpha}{greek}{"22}
  \DeclareMathSymbol{\vartheta}{\mathalpha}{greek}{"23}
  \DeclareMathSymbol{\varpi}{\mathalpha}{greek}{"24}
  \DeclareMathSymbol{\varrho}{\mathalpha}{greek}{"25}
  \DeclareMathSymbol{\varsigma}{\mathalpha}{greek}{"26}
  \DeclareMathSymbol{\varphi}{\mathalpha}{greek}{"27}
  \DeclareSymbolFont{otone}{OT1}{cmr}{m}{n}
  \DeclareMathSymbol{\Gamma}{\mathalpha}{otone}{0}
  \DeclareMathSymbol{\Delta}{\mathalpha}{otone}{1}
  \DeclareMathSymbol{\Theta}{\mathalpha}{otone}{2}
  \DeclareMathSymbol{\Lambda}{\mathalpha}{otone}{3}
  \DeclareMathSymbol{\Xi}{\mathalpha}{otone}{4}
  \DeclareMathSymbol{\Pi}{\mathalpha}{otone}{5}
  \DeclareMathSymbol{\Sigma}{\mathalpha}{otone}{6}
  \DeclareMathSymbol{\Upsilon}{\mathalpha}{otone}{7}
  \DeclareMathSymbol{\Phi}{\mathalpha}{otone}{8}
  \DeclareMathSymbol{\Psi}{\mathalpha}{otone}{9}
  \DeclareMathSymbol{\Omega}{\mathalpha}{otone}{10}
  \DeclareSymbolFont{syms}{OML}{cmm}{m}{it}
  \DeclareMathSymbol{\partial}{\mathord}{syms}{"40}
  \DeclareMathAlphabet{\mathbold}{OML}{cmm}{b}{it}
  \DeclareSymbolFont{largesymbols}{OMX}{cmex}{m}{n}
